%% file: iflytek_paper.tex
\documentclass[]{iflytek_robotics}

\input{macros}
\usepackage{array}
\usepackage{xcolor}

\title{EffectWAM: Learning Action-Grounded Effects for Robot Manipulation}

\author[1,\dagger]{Yuan Zhang}
\author[1,\dagger]{Hao Cui}
\author[1,\dagger]{Jingfei Ni}
\author[1]{Chi Liu}
\author[1]{Xin Nie}
\author[1]{Jiajia Wu}
\author[1]{Meiyi Wang}
\author[1]{Wenjie Xu}
\author[1]{Lin Gao}
\author[1]{Hao Li}
\author[1]{Xianhe Zhang}
\author[2]{Diyuan Liu}
\author[1,*]{Jia Pan}
\author[2,*]{Chao Ji}

\affiliation[1]{iFLYTEK}
\affiliation[2]{LindenBot}

\abstract{
World--action models connect visual prediction with executable robot control. We present EffectWAM, which develops this connection through three complementary components: Effect, camera positional encoding (CaPE), and Action Grounding. Effect represents visual change as the difference between independently encoded future observations and a shared current-observation anchor. CaPE conditions endpoint pooling on relative camera pose when reliable extrinsics are available, while sharing the same learned parameters with pose-free pooling for uncalibrated trajectories. Action Grounding trains the visual representation to retain measured changes in end-effector pose and gripper state. An exponential moving average of the observation encoder supplies Effect targets. A shared generation expert learns to jointly predict Effect and an executable action chunk through flow matching, and reuses the sampled pair to generate future video on demand. This allows the controller to execute actions without waiting for video synthesis. We train on more than 20,000 hours of heterogeneous embodied data, and accommodate dataset-specific joint, end-effector, and hand-pose representations. Our evaluation uses the Small (8B) configuration on all five benchmarks and additionally evaluates the Large (26B) configuration on RoboDojo. Experiments examine closed-loop manipulation across complementary benchmarks, and comparisons across multiple baseline policies show consistent improvements from Action Grounding.
}

\date{November 27, 2026}
\correspondence{\email{jiapan@iflytek.com}; \email{chaoji@iflytek.com}}
\checkdata[Project Code]{\url{https://github.com/haocui1715/xx-wam-mem-0.5-preview}}
\checkdata[Project Model]{\url{https://huggingface.co/haocui/xx-wam-mem-0.5-preview}}

\begin{document}
\bstctlcite{IEEEBSTcontrol}
\maketitle

\input{iFlyBot_sections/introduction}
\input{iFlyBot_sections/related_work}
\input{iFlyBot_sections/model}
\input{iFlyBot_sections/training_strategy}
\input{iFlyBot_sections/experiments}
\input{iFlyBot_sections/conclusions}


\bibliographystyle{IEEEtran}
\bibliography{references}

\clearpage
\input{iFlyBot_sections/appendix_results}

\end{document}

%% file: macros.tex
\usepackage{minitoc}
\usepackage{url}
\usepackage{amssymb}
\usepackage[utf8]{inputenc}
\usepackage{microtype}
\usepackage{booktabs}
\usepackage{pifont} 
\usepackage{multirow}
\usepackage{makecell}
\usepackage{paralist}
\usepackage{xspace}
\usepackage{color}
\usepackage{xcolor}
\usepackage{colortbl}
\usepackage{adjustbox}
\usepackage{hyperref} 
\usepackage[edges]{forest}
\usepackage{tikz} 
\usepackage{caption}
\usepackage{amsfonts}
\usepackage[toc,page,header]{appendix}
\usepackage[utf8]{inputenc}
\usepackage{bbm}
\usepackage{enumitem}

\definecolor{seedc}{RGB}{7, 92, 173}

\newcommand{\name}[1]{EffectWAM}
\renewcommand{\paragraph}[1]{\vspace{0.1em}\noindent\textbf{#1}}

%% file: iFlyBot_sections/introduction.tex
\section{Introduction}
\label{sect:intro}

Language-conditioned robot manipulation requires a policy to translate a task instruction into a sequence of actions that brings about the intended change in the environment. In closed-loop execution, the policy repeatedly observes the scene and adjusts its actions as the task progresses. Anticipating how the scene may evolve over an action horizon can help the policy connect its immediate commands to the intended outcome. The choice of what to predict about the future is therefore central to learning visual dynamics for robot control.

Vision--language--action (VLA) models learn to map observations and instructions to robot actions \cite{kim2024openvlaopensourcevisionlanguageactionmodel,black2026pi0visionlanguageactionflowmodel}. Their action supervision leaves the visual consequences of those commands implicit. World--action models (WAMs) add an explicit objective for predicting visual dynamics \cite{ye2026worldactionmodelszeroshot,miao2026internw0delta}. Future video provides a rich training signal, but synthesizing it at every decision adds latency to closed-loop control. Fast-WAM avoids this cost by using future video only during training~\cite{yuan2026fastwam}. In matched experiments, LAWA found that this approach generalized less well than joint video and action prediction, especially with limited robot demonstrations, and showed that predicting a compact latent alongside actions could achieve comparable success to joint video and action prediction on RoboCasa at lower latency~\cite{li2026lawa}. This motivates a policy that retains a compact prediction of action consequences at test time to support manipulation performance while keeping video synthesis off the control path.

The choice of target determines which changes the policy learns to anticipate. La-JEPA, for example, uses a frozen V-JEPA 2.1 encoder to supervise future prediction alongside robot actions~\cite{murlabadia2026vjepa21,clemente2026lajepa}. Frozen video features inherit the distinctions learned during pretraining. A generic video prediction objective provides no direct supervision for retaining small changes in gripper opening or end-effector pose, even when those changes matter for control. Robot demonstrations provide that signal through synchronized images and measured state changes. We use these measurements to shape the representation that supplies the policy's future prediction targets.

To this end, we introduce Effect, a compact representation of visual change over the action horizon that learns from paired observations and robot measurements in the same demonstration. A shared encoder processes the current and selected future observations independently. We subtract the current feature from each future feature so that changes at different horizons are expressed relative to the state from which the robot acts. An auxiliary head reads out measured changes in end-effector pose and gripper state from the feature difference, encouraging the encoder to preserve information about the executed motion. An exponential moving average (EMA) of the encoder supplies the resulting Effect targets during policy training.

Across robot embodiments and camera setups, similar manipulation motions can yield different image changes because cameras are mounted and move differently. For example, as a wrist camera approaches a drawer, the handle can shift in the image before the drawer moves. Qwen-RobotManip uses explicit camera geometry to align visual observations and end-effector motion across heterogeneous robot data~\cite{yuan2026qwenrobotmanip}. To support transfer across embodiments and camera setups, we condition Effect's attention pooling on the camera pose relative to the current observation when reliable extrinsics are available. This gives the encoder geometric context for distinguishing changes due to camera motion from changes in the scene. When extrinsics are unavailable, the same encoder uses pose-free pooling with shared weights, allowing both types of trajectory to contribute to training.

EffectWAM integrates Effect, CaPE, and Action Grounding through two complementary flow-matching objectives. The first jointly models Effect and the demonstrated action chunk given the current observation and instruction. The second models future video latents given the same context, the clean Effect target, and the demonstrated action. At inference, the policy jointly samples Effect and the action chunk, retaining a prediction of visual change in each control decision. The controller can execute the sampled chunk as soon as joint sampling finishes. When a visual forecast is requested, the video mode uses the same Effect--action pair.

We train with more than 20,000 hours of embodied data spanning heterogeneous action representations. We evaluate the Small (8B) configuration on LIBERO~\cite{liu2023libero}, LIBERO-Plus~\cite{fei2026liberoplus}, RoboTwin 2.0~\cite{chen2026robotwin}, RoboCasa365~\cite{nasiriany2026robocasa365}, and RoboDojo~\cite{chen2026robodojo}, and additionally evaluate the Large (26B) configuration on RoboDojo. The experimental comparisons distinguish these configurations and compare Small (8B) with existing policies. We also examine Action Grounding across multiple baseline models, where it consistently improves performance, supporting its value beyond a single architecture.

Our contributions are as follows.
\begin{itemize}
    \item \textbf{Effect: a compact prediction of visual change.} We represent future observations relative to a shared current-observation anchor and jointly predict the resulting Effect sequence with executable action chunks. The same Effect--action pair conditions optional video generation, connecting control to visual forecasting without requiring video synthesis at each decision.

    \item \textbf{CaPE: geometry-aware visual transition encoding.} We condition endpoint pooling on relative camera pose to account for viewpoint changes across robot and camera configurations. Calibrated and uncalibrated trajectories share the same learned pooling parameters, supporting joint learning from heterogeneous demonstrations.

    \item \textbf{Action Grounding: supervision from measured motion.} We make end-effector and gripper changes recoverable from Effect, shaping the visual representation around motion relevant to robot control. Evaluations across multiple baseline policies show consistent improvements from this objective, indicating that its benefit extends beyond our complete model.
\end{itemize}

%% file: iFlyBot_sections/related_work.tex
\section{Related Work}

\paragraph{Predictive targets for world-action models.}
Future video provides a learning signal for control, but different models use that signal at different points in the policy. Video-first policies infer controls from predicted frames \cite{du2023learninguniversalpoliciestextguided}, while DreamZero jointly generates video and actions \cite{ye2026worldactionmodelszeroshot}. Fast-WAM retains video prediction during training and skips it during action inference \cite{yuan2026fastwam}. Predicting in a compact feature space offers another way to use future observations. V-JEPA 2.1 learns dense spatiotemporal features through self-supervised video pretraining \cite{murlabadia2026vjepa21}. La-JEPA uses a pretrained V-JEPA 2.1 encoder to define the future features that its policy predicts alongside actions during training \cite{clemente2026lajepa}. InternW0-$\Delta$ uses changes in video features more explicitly. Its Causal Imprint regresses differences between adjacent video latents and aligns intermediate features with future-video features before conditioning a separate action expert \cite{miao2026internw0delta}. In each case, the feature space used to construct the prediction targets shapes which visual changes the policy is trained to anticipate.

\paragraph{Learning representations of visual transitions.}
Latent-action models learn a representation of the transition itself. LAPA compresses changes between frame pairs into discrete codes that support future-frame reconstruction and uses those codes for policy pretraining \cite{ye2025lapa}. Subsequent work uses additional signals to capture manipulation-relevant information. WALA learns latent actions by predicting current-anchored changes in pretrained semantic features and estimated depth, then uses its pretrained latent-action encoder to supervise the policy \cite{liu2026wala}. LAWA uses a tokenizer trained with next-frame reconstruction and manipulation-region masks, then jointly denoises a continuous latent and executable actions \cite{li2026lawa}. Robot measurements can also shape the representation. WLA$^3$ encodes synchronized views together with available embodiment-state changes and learns through multimodal reconstruction and consistency across overlapping windows \cite{liu2026wla3}. In our method, end-effector and gripper measurements supervise an auxiliary readout from visual feature differences. The observation encoder learns during policy training, and its EMA copy supplies the Effect targets. The resulting Effect is jointly predicted with actions and can also condition a video forecast together with the sampled action chunk. We treat Action Grounding as a distinct supervision mechanism and evaluate its contribution across multiple baseline policies, in addition to evaluating the complete model.

\paragraph{Camera geometry in visual transition encoding.}
A visual transition can reflect both physical interaction and a change in viewpoint. EscherNet introduces camera positional encoding (CaPE) to express relative camera pose in attention, and PRoPE extends relative camera encoding to include camera intrinsics and extrinsics \cite{kong2024eschernet,li2025camerasrelative}. We use CaPE to provide the Effect encoder with explicit viewpoint information when it pools each endpoint. We express available camera poses relative to the current observation and use the same pooling weights for trajectories with and without reliable extrinsics.

%% file: iFlyBot_sections/model.tex
\section{Model Design}
\label{sect:model}

\subsection{Problem Formulation}
\label{sec:problem_formulation}

At each decision time $t_0$, the policy observes the available RGB camera views $I_0^{1:V}$ and receives a language instruction $\ell$. We denote the conditioning information by $C=(I_0^{1:V},\ell,\mathcal S,\mathcal Q)$, where $\mathcal S$ specifies the robot's action representation and $\mathcal Q$ specifies the prediction horizons relative to $t_0$. The model predicts an executable action chunk $A=(a_1,\ldots,a_{N_A})$ and an Effect sequence $E=(e_1,\ldots,e_{N_E})$. Each $e_k$ represents the change from the current observation to a requested future time. Actions and Effects are indexed by their respective temporal supports. We denote the latent of the future video by $Y$.

We model the three variables through the conditional factorization
\begin{equation}
    p_\theta(E,A,Y\mid C)
    = p_\theta(E,A\mid C)\,p_\theta(Y\mid C,E,A).
    \label{eq:world_action_factorization}
\end{equation}
The first factor gives the policy an action chunk and a prediction of the accompanying visual change. The second uses that pair to produce a visual forecast on request. During training, future observations provide the targets for $E$ and $Y$. The policy conditions its joint Effect--action prediction on the current information in $C$.

\subsection{Architecture Overview}
\label{sect:overview}
\label{sec:architecture_overview}

\begin{figure}[tbp]
    \centering
    \includegraphics[width=\linewidth]{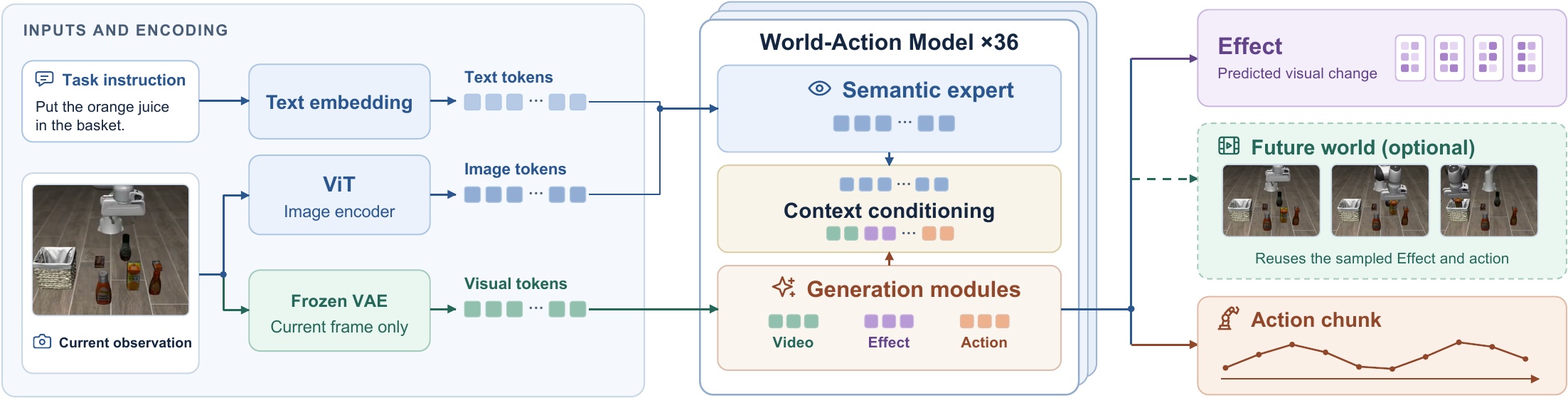}
    \caption{World--action inference. The World--Action mixture-of-transformers (MoT) combines a semantic expert (SEM) and a shared generation expert (GEN) through masked self-attention. SEM receives instruction and image tokens, and GEN receives visual tokens from the current observation encoded by a frozen video VAE. GEN jointly samples an Effect sequence $\hat E$ and an action chunk $\hat A$. On request, a second generation pass uses the same sampled pair to produce a future video.}
    \label{fig:framework}
\end{figure}

EffectWAM builds a common observation--language context for both factors in Eq.~\eqref{eq:world_action_factorization}, linking joint Effect--action prediction with conditional video generation (Fig.~\ref{fig:framework}). The task instruction is embedded as text tokens, and a trainable vision transformer (ViT) maps the available current camera views to merged image tokens. SEM receives both sets of tokens, with intermediate visual features injected into early SEM blocks. In parallel, the frozen video VAE encodes the current observation alone. A learned projection maps its latent to visual tokens for GEN.

The mixture-of-transformers (MoT) uses masked self-attention to couple SEM and GEN. GEN shares its transformer backbone across Effect, action, and video, with modality-specific input and output projections. It first samples the Effect and action streams jointly. The optional video pass reuses the sampled pair and the original observation context, as detailed in Sec.~\ref{sec:closed_loop_inference}. During training, the ViT also supplies pre-merger patch features to the observation endpoint encoder in Sec.~\ref{sec:effect_learning}. The component parameter counts for the Small (8B) and Large (26B) configurations are reported in Table~\ref{tab:model_configurations}.

\subsection{Physically Grounded Effect Learning}
\label{sec:effect_learning}
\label{sec:camera_pose_pooling}
\label{sec:action_grounding}

\begin{figure}[tbp]
    \centering
    \includegraphics[width=\linewidth]{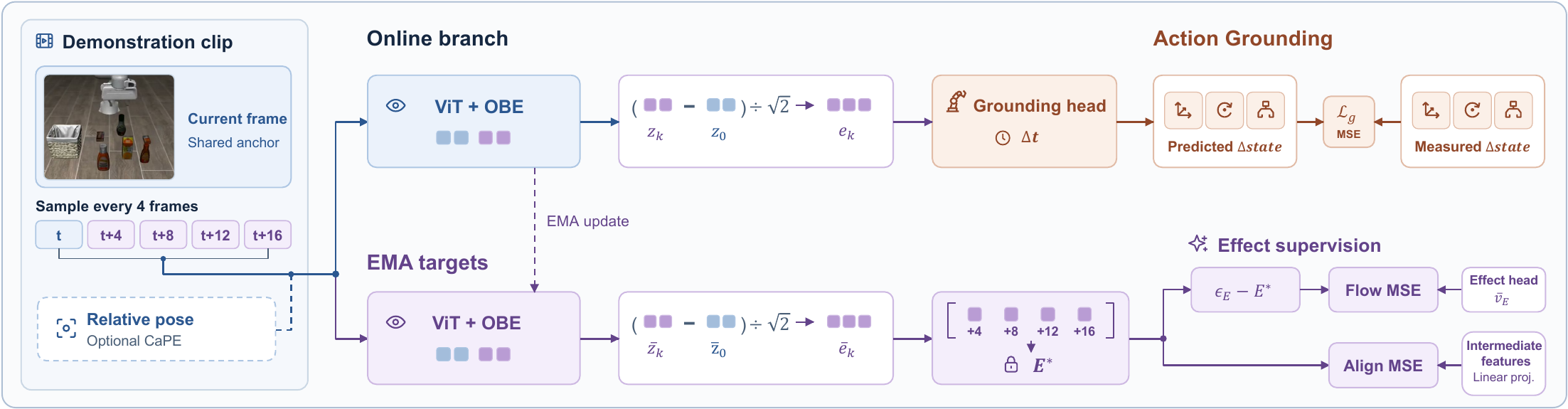}
    \caption{Physically grounded Effect learning. The online and exponential moving average (EMA) branches encode the same demonstration frames and form endpoint differences relative to the shared current frame. Optional camera positional encoding (CaPE) conditions both branches. The online grounding head uses $e_k$ and the elapsed interval $\Delta t_k$ to predict measured state changes. The EMA branch stacks its differences into the detached target $E^\star$. The final GEN head is supervised with the velocity target $\epsilon_E-E^\star$, and a linear projection of the intermediate Effect states (block~9 in the reference configuration) is supervised with $E^\star$ directly.}
    \label{fig:effect_grounding}
\end{figure}

\paragraph{Effect representation.}
We learn Effect from a primary-camera demonstration clip by expressing future observations relative to a shared current-observation anchor (Fig.~\ref{fig:effect_grounding}). The reference configuration samples future frames at offsets of $4$, $8$, $12$, and $16$ frames. The ViT encodes each sampled frame independently. Given the pre-merger patch features of frame $I_k$, the observation endpoint encoder (OBE) pools the patches using one learned query and four attention heads. An output projection and a two-layer MLP produce $u_k\in\mathbb R^d$, where $d=256$. The radial map $z_k=\sqrt{d}\,u_k/\sqrt{d+\lVert u_k\rVert_2^2}$ gives a bounded endpoint feature. Each Effect token compares a future endpoint with the same current endpoint $z_0$,
\begin{equation}
    e_k=\frac{z_k-z_0}{\sqrt{2}}.
    \label{eq:effect_difference}
\end{equation}
The sequence of differences describes changes from the state at which the action chunk is issued to each requested future time.

\paragraph{Camera pose-aware pooling (CaPE).}
These endpoint differences can reflect camera motion as well as physical interaction. We incorporate camera positional encoding (CaPE)~\cite{kong2024eschernet} into endpoint pooling when reliable extrinsics are available. Let $T_{C_k\leftarrow W}$ denote the world-to-camera transform for the primary camera at time $t_k$. We express pose relative to the current camera,
\begin{equation}
    P_k=T_{C_k\leftarrow W}\bigl(T_{C_0\leftarrow W}\bigr)^{-1},
    \qquad P_0=I_4.
    \label{eq:relative_camera_pose}
\end{equation}
For a projected attention key $k_j$ at patch $j$ of frame $k$, we split each key head into four-dimensional row-vector blocks and right-multiply each block by $P_k$. Equivalently, $k'_j=k_j\operatorname{diag}(P_k,\ldots,P_k)$. All patches in a frame share the same relative transform. The query and values retain their original projections. For trajectories without camera pose conditioning, pooling uses the untransformed keys. The online and EMA branches use the same relative poses when available. Within each branch, both pooling routes use the same learned query, projections, and endpoint mapping, and they coincide when $P_k=I_4$.

\paragraph{Action Grounding and EMA targets.}
To relate these visual differences to measured physical changes, the grounding head takes the online Effect $e_k$ and the elapsed interval $\Delta t_k$ from the current frame to the corresponding future frame. It predicts changes in end-effector translation, rotation, and gripper state over that interval. The physical grounding loss $\mathcal L_g$ compares these predictions with measurements using squared error over available quantities after normalization within each measurement family. We also regularize the current endpoint using consistency across two photometric views of the same observation and constraints on feature variance and covariance.

To construct prediction targets from the learned endpoint representation, an exponential moving average (EMA) tracks the pre-merger ViT and the OBE. The EMA branch encodes the same frames and computes endpoint differences relative to its own current endpoint $\bar z_0$. With $\bar z_k$ denoting the future EMA endpoint and $\operatorname{sg}$ the stop-gradient operator, we stack the differences in temporal order to obtain the clean target,
\begin{equation}
    e_k^\star=\operatorname{sg}\!\left(\frac{\bar z_k-\bar z_0}{\sqrt{2}}\right),
    \qquad E^\star=(e_1^\star,\ldots,e_{N_E}^\star).
    \label{eq:effect_target}
\end{equation}
As shown in Fig.~\ref{fig:effect_grounding}, $E^\star$ supplies both the Effect flow-matching target and the intermediate alignment target. It also conditions the video mode during training, as detailed in Sec.~\ref{sec:shared_generation}.

\subsection{Shared Generation and Training Objectives}
\label{sec:shared_generation}

We implement the two factors of Eq.~\eqref{eq:world_action_factorization} as separate forward modes of GEN with shared parameters. The joint Effect--action mode (\texttt{ACT\_EA}) takes noisy Effect and action tokens and predicts their velocities using the current context. The video mode (\texttt{VIDEO\_EA}) takes noisy future video tokens, conditioned on the same context and a clean Effect--action pair. During training, this pair consists of the detached EMA Effect target $E^\star$ and the demonstrated action chunk $A^\star$. Table~\ref{tab:prediction_modes} summarizes the two modes.

\begin{table}[tbp]
    \centering
    \caption{Generation modes during training. Both modes receive the current context $C$.}
    \label{tab:prediction_modes}
    \begin{tabular}{@{}lcc@{}}
        \hline
        Mode & Additional clean conditions & Noisy prediction variables \\
        \hline
        \texttt{ACT\_EA} & None & $E_\sigma, A_\sigma$ \\
        \texttt{VIDEO\_EA} & $E^\star, A^\star$ & $Y_\sigma$ \\
        \hline
    \end{tabular}
\end{table}

Within a sample, semantic tokens attend causally within their own stream. Generation tokens attend to the semantic tokens and to the clean and noisy tokens of the active generation mode. The diffusion timestep embedding is applied to the noisy tokens. The two modes enter the same optimization window through separate forward passes. The joint Effect--action mode uses the observed context and its own noisy variables. The frozen video VAE encodes the consecutive demonstration clip to obtain the future video target $Y^\star$. In the reference configuration, its $4\times$ temporal compression places the future latent slices at the same frame offsets as the Effect endpoints in Fig.~\ref{fig:effect_grounding}. The visual condition in $C$ comes from a separate encoding of the current observation alone.

For each predicted variable $x\in\{E,A,Y\}$, we draw Gaussian noise $\epsilon_x\sim\mathcal N(0,I)$ and, in the reference configuration, a flow time $\sigma=5u/(1+4u)$ with $u\sim\mathcal U(0,1)$. We interpolate between the clean target $x^\star$ and noise and regress the velocity along that path,
\begin{equation}
    x_\sigma=(1-\sigma)x^\star+\sigma\epsilon_x,
    \qquad v_x^\star=\epsilon_x-x^\star.
    \label{eq:effect_flow_matching}
\end{equation}
Effect and action share the flow time and receive independent Gaussian noise. In video mode only the future video latent is noised. We define $\mathcal L_e$, $\mathcal L_a$, and $\mathcal L_y$ as the corresponding squared velocity errors, averaged over valid coordinates and then over eligible requests. In Fig.~\ref{fig:effect_grounding}, the final GEN head predicts $\hat v_E$, which is compared with the velocity target $\epsilon_E-E^\star$.

In the reference configuration, the intermediate alignment loss $\mathcal L_{\mathrm{align}}$ applies a learned linear projection to the GEN block~9 hidden states at the Effect-token positions and uses squared error against the clean target $E^\star$. This loss is applied to eligible noisy Effect tokens with $\sigma\geq 0.5$ once enabled according to the training schedule. The observation loss $\mathcal L_{\mathrm{obs}}$ matches the online endpoint from each photometric view to the detached EMA endpoint of the other view. It also penalizes low per-dimension variance and off-diagonal covariance within compatible groups. Together with the grounding loss in Sec.~\ref{sec:action_grounding}, the combined objective is
\begin{equation}
    \begin{aligned}
    \mathcal L={}&\lambda_e\mathcal L_e+\lambda_a\mathcal L_a+\lambda_y\mathcal L_y
    +\lambda_g\mathcal L_g \\
    &+\lambda_{\mathrm{obs}}\mathcal L_{\mathrm{obs}}
    +\lambda_{\mathrm{align}}\mathcal L_{\mathrm{align}}.
    \end{aligned}
    \label{eq:combined_training_objective}
\end{equation}

\subsection{Closed-Loop Inference and On-Demand Video Generation}
\label{sec:closed_loop_inference}

At each decision time, the policy constructs $C$ from the latest RGB observations, the task instruction, the action interface, and the requested prediction horizons. The ViT-derived semantic features and the frozen VAE latent of the current observation provide the visual context for sampling. The observation endpoint encoder, physical readout, and EMA target encoder are used during training. At inference, GEN samples the Effect sequence and action chunk jointly from the current context, as shown in Fig.~\ref{fig:framework}.

In the joint Effect--action mode, we initialize the Effect and action variables with independent Gaussian noise and treat them as one state $s_\sigma=(E_\sigma,A_\sigma)$ on a shared flow time grid. At each solver evaluation, the generation expert reads both noisy streams and predicts their velocities together,
\begin{equation}
    \frac{\mathrm d s_\sigma}{\mathrm d\sigma}
    =\bigl(\hat v_E(s_\sigma,\sigma,C),\hat v_A(s_\sigma,\sigma,C)\bigr).
    \label{eq:joint_inference_flow}
\end{equation}
We integrate toward $\sigma=0$ to obtain $(\hat E,\hat A)$. The reference configuration uses a second-order Flow UniPC solver with four integration steps per generation mode. With classifier-free guidance, the conditional and unconditional velocity estimates are evaluated on the same noisy state and combined before each solver update. Generation-token hidden states are recomputed at each evaluation because they can attend to the evolving noisy variables.

For closed-loop control, we denormalize the valid coordinates of $\hat A$ and convert them to the command representation specified by $\mathcal S$. The controller executes a configured prefix of the chunk, which can be the entire chunk, then acquires a new observation and constructs the next context. The next decision uses the new context to sample a fresh Effect--action pair. The joint sampler therefore supplies the commands needed for execution as soon as its integration completes.

When a visual forecast is requested, the video mode retains the observation snapshot and prediction horizons used to generate $(\hat E,\hat A)$. It takes this exact pair as clean conditions in the generator's representation and starts from an independent video noise sample. The same generation expert integrates the video velocity to obtain $\hat Y$, and the frozen VAE decodes the current and predicted future latents into RGB frames. The forecast describes the sampled action chunk from its original observation, including the planned suffix if the controller executes only a prefix before replanning. This procedure keeps video generation optional and ties the forecast to the same action proposal used for control.

%% file: iFlyBot_sections/training_strategy.tex
\section{Training Strategy}
\label{sect:strategy}

\subsection{Data Preparation}
\label{sec:data_preparation}
\label{sec:action_representations}

EffectWAM learns to predict visual changes together with executable actions. Training therefore combines robot demonstrations, human manipulation recordings, and egocentric video, which provide complementary coverage of robot control, hand--object interaction, and visual dynamics. The mixture spans different embodiments, camera configurations, manipulation skills, and levels of action annotation.

\begin{figure}[!htbp]
    \centering
    \includegraphics[width=\linewidth]{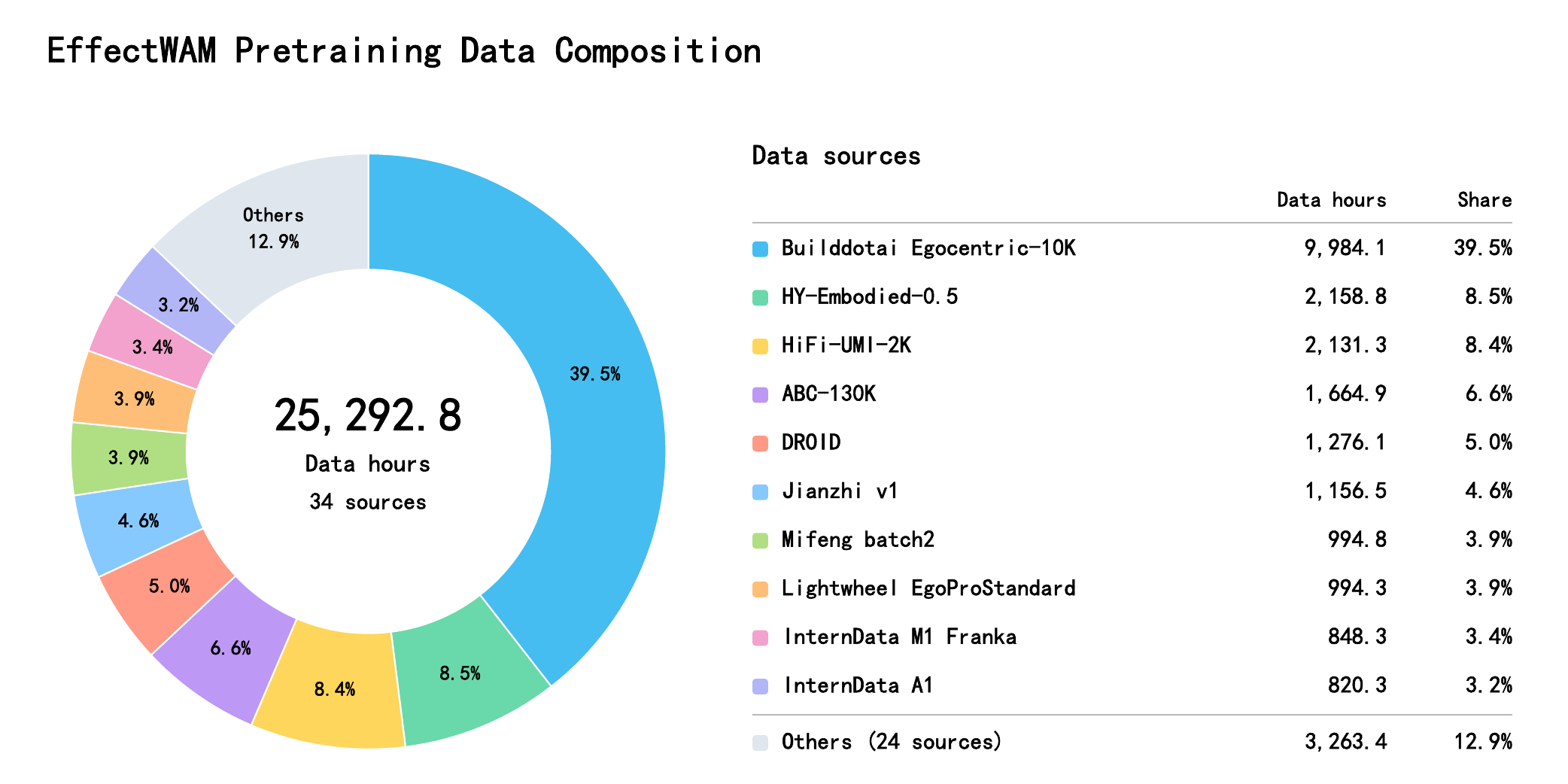}
    \caption{EffectWAM pretraining data composition by recorded duration. The donut chart shows the ten largest sources individually and groups the remaining 24 as Others (3,263.4 hours; 12.9\%). Data hours denotes recorded episode duration, with all shares normalized by the 25,292.8-hour total.}
    \label{fig:dataset_composition}
\end{figure}

We pretrain on more than 20,000 hours of embodied interaction data. The inventory contains 25,292.8 hours across 34 sources. Figure~\ref{fig:dataset_composition} summarizes this inventory by recorded episode duration; the displayed shares are not training sampling weights. Appendix~\ref{app:data_inventory} provides the complete source list.

The largest sources include Builddotai Egocentric-10K, HY-Embodied-0.5, HiFi-UMI-2K, ABC-130K, and DROID. The remaining mixture includes InternData, RH20T, EgoDex, RoboCOIN, RoboMIND, FurnitureBench, and additional human and robot interaction collections. Others also includes the official RoboDojo simulation training dataset (20.5 hours)~\cite{chen2026robodojo} and GPT-Astra robot-operation data (10.5 hours).

Robot demonstrations provide synchronized visual observations and control annotations for action learning. Their original records may include joint positions, end-effector poses, and gripper states or commands; dexterous-robot demonstrations may also provide finger-joint or hand-pose coordinates. We preserve the physical meaning of these annotations through the corresponding action interfaces.

Human manipulation recordings broaden the observed range of object interactions and hand motions. Some recordings include tracked wrist and articulated hand poses, which describe human motion rather than executable robot commands. Video-only recordings contribute temporal visual supervision where action or physical-state labels are absent. Table~\ref{tab:action_representations} summarizes representative record types and their original annotations. These differences in annotation determine the supervision available during training.

\begin{table}[!htbp]
\centering
\caption{Original action/state annotations across representative record types. Available fields vary by source and processing version.}
\label{tab:action_representations}
\small
\begin{tabular}{@{}>{\raggedright\arraybackslash}p{0.34\linewidth}>{\raggedright\arraybackslash}p{\dimexpr0.66\linewidth-2\tabcolsep\relax}@{}}
\toprule
Data source / Record type & Original action / State annotations \\
\midrule
Robot arm and gripper trajectories & Joint positions, end-effector translation and orientation, and gripper states or commands, where provided. \\
Dexterous robot demonstrations & Arm pose plus finger-joint or hand-pose coordinates. \\
Human hand recordings & Tracked wrist and articulated hand poses. \\
Video-only egocentric recordings & No action or physical-state annotations; RGB sequences and available task text. \\
\bottomrule
\end{tabular}
\end{table}

\subsection{Training Process}
\label{sec:training_process}

Training follows the objectives in Sec.~\ref{sec:shared_generation}, combining the supervision available in each trajectory. The current camera observations, task instruction, action interface, and prediction horizons define the conditioning context. The trainable ViT supplies visual features to the semantic expert, while the frozen VAE separately encodes the current observation to provide the observed visual latents.

The prediction targets are constructed from the same demonstration. The EMA branch encodes the current and selected future observations independently through the pre-merger ViT and observation endpoint encoder. Future endpoints are compared with their shared current endpoint to form detached Effect targets, as defined in Eq.~\eqref{eq:effect_target}. The frozen VAE encodes the demonstration clip to supply future-video targets. The current visual context remains distinct from these future prediction targets.

The shared generation expert is trained through two forward modes within the same optimization window. Joint Effect--action training uses a shared flow time and independent noise for Effect and actions. Video training noises the future-video latent while conditioning on the clean EMA Effect target and demonstrated action chunk. Both modes share generation parameters and optimize their corresponding velocity-prediction objectives.

Each mode requires the annotations and conditioning variables specified in Table~\ref{tab:prediction_modes}. Available physical measurements supervise Action Grounding through changes in end-effector pose and gripper state over the prediction interval. Reliable camera extrinsics enable CaPE-conditioned endpoint pooling; uncalibrated trajectories use the pose-free route with the same learned pooling parameters. Observation consistency and intermediate Effect alignment provide the additional supervision defined in Sec.~\ref{sec:shared_generation}. Terms requiring unavailable annotations are excluded. Visual-only recordings contribute the applicable visual objectives without supplying robot-action labels.

Following training on the multi-source data, the model adapts to each benchmark's demonstrations using the corresponding action interface. This preserves the source-specific meaning of action annotations while applying the shared training objectives.

%% file: iFlyBot_sections/experiments.tex
\section{Experiments}
\label{sect:experiments}

We evaluate EffectWAM on five manipulation benchmarks to study task execution, robustness, compositional generalization, and capability-specific behavior. The comparisons examine task success relative to InternW0-$\Delta$~\cite{miao2026internw0delta}, robustness to environmental perturbations, and generalization to unfamiliar task combinations.

\subsection{Experimental Setup}
\label{sec:model_scale}

We evaluate two configurations, \textbf{Small (8B)} and \textbf{Large (26B)}, with component sizes listed in Table~\ref{tab:model_configurations}. Small is evaluated on all five benchmarks, and Large is additionally evaluated on RoboDojo. The two RoboDojo configurations are compared with baselines of similar model size in separate tables.

\begin{table}[tbp]
\centering
\caption{Model configurations of EffectWAM Small (8B) and Large (26B). Parameter counts refer to the listed modules.}
\label{tab:model_configurations}
\small
\setlength{\tabcolsep}{5pt}
\begin{adjustbox}{max width=\linewidth}
\begin{tabular}{@{}llc@{}}
\toprule
Scale & Module & Parameters (B) \\
\midrule
\multirow{3}{*}{Small (8B)} & VLM & 2.2 \\
 & Video DiT & 5.0 \\
 & Action DiT & 1.2 \\
\midrule
\multirow{3}{*}{Large (26B)} & VLM & 8.3 \\
 & Video DiT & 16.4 \\
 & Action DiT & 2.0 \\
\bottomrule
\end{tabular}
\end{adjustbox}
\end{table}

Section~\ref{sec:task_performance} evaluates standard task execution and robustness to changes in observations, robot initial states, and scene configuration. LIBERO measures task performance relative to InternW0-$\Delta$, while LIBERO-Plus and the Clean/Randomized comparison on RoboTwin 2.0 assess performance under distribution shifts. Section~\ref{sec:compositional_manipulation} examines whether the gains extend beyond atomic tasks to seen and unseen combinations on RoboCasa365. RoboDojo complements these comparisons with a capability-level analysis, using success rates and partial-credit task scores to characterize full-task completion and intermediate progress across five capability dimensions.

On LIBERO and RoboCasa365, we evaluate InternW0-$\Delta$ (8B) and EffectWAM using the same task sets and episode budgets. EffectWAM uses only Clean demonstrations for RoboTwin 2.0 and only Human300 for RoboCasa365, without Target training data. The RoboTwin comparison is restricted to policies adapted using only Clean demonstrations from this benchmark. Model sizes are approximate and include frozen conditioning encoders but exclude VAEs.

We follow each benchmark's aggregation protocol. LIBERO-Plus and RoboCasa365 aggregate success over evaluation episodes, RoboTwin 2.0 weights Clean and Randomized success equally, and RoboDojo weights its five capability dimensions equally. Success rates are reported as percentages; RoboDojo additionally reports partial-credit task scores; Small is evaluated with one seed, while Large reports variation across three seeds.

\subsection{Task Performance and Robustness}
\label{sec:task_performance}
\label{sec:libero_plus}
\label{sec:robotwin}

\paragraph{Standard task execution.}
We use standard LIBERO to establish task execution performance before introducing distribution shifts. We evaluate EffectWAM and InternW0-$\Delta$ across four suites with 10 tasks per suite and 50 episodes per task, totaling 2,000 episodes per model. EffectWAM Small achieves 99.20\% overall success, compared with 99.00\% for InternW0-$\Delta$ (Table~\ref{tab:libero_standard}). The comparison spans supervised policies and methods with additional post-training, providing a broader reference for standard task execution.

\begin{table}[tbp]
\centering
\caption[Standard LIBERO success rates (\%).]{%
Standard LIBERO success rates (\%).
The highest score is shown in
\textcolor{red}{\textbf{bold red}}, and the second-highest distinct score
is \underline{underlined}.}
\label{tab:libero_standard}
\small
\setlength{\tabcolsep}{4pt}
\begin{adjustbox}{max width=\linewidth}
\begin{tabular}{@{}lcccccc@{}}
\toprule
Method & Size (B) & Spatial & Object & Goal & Long & Overall \\
\midrule
OpenVLA-OFT (joint)~\cite{kim2025fine} & 7.0 & 97.7 & 98.0 & 96.1 & 95.3 & 96.8 \\
UniVLA~\cite{bu2025univla} & 7.0 & 96.5 & 96.8 & 95.6 & 92.0 & 95.2 \\
RIPT-VLA~\cite{tan2025riptvla} & 7.0 & 98.6 & 98.6 & 99.0 & 93.8 & 97.5 \\
NORA-1.5 (DPO)~\cite{hung2025nora15} & 4.0 & 98.0 & 96.0 & 95.4 & 90.5 & 95.0 \\
SRPO~\cite{fei2025srpo} & 7.0 & \underline{98.8} & \textcolor{red}{\textbf{100}} & \textcolor{red}{\textbf{99.4}} & \underline{98.6} & \textcolor{red}{\textbf{99.2}} \\
AVA-VLA (joint)~\cite{xiao2026avavla} & 7.0 & 97.4 & 99.4 & 97.4 & 97.6 & 98.0 \\
InternW0-$\Delta$~\cite{miao2026internw0delta} & 8.0 & \textcolor{red}{\textbf{99.2}} & \underline{99.8} & 98.0 & \textcolor{red}{\textbf{99.0}} & \underline{99.0} \\
\textbf{EffectWAM Small} & 8.0 & \textcolor{red}{\textbf{99.2}} & \textcolor{red}{\textbf{100}} & \underline{99.2} & 98.4 & \textcolor{red}{\textbf{99.2}} \\
\bottomrule
\end{tabular}
\end{adjustbox}
\end{table}

EffectWAM improves overall success from an already high baseline, with the clearest gain on the Goal suite: success rises from 98.00\% to 99.20\%. Object success also increases to 100.00\%, while Spatial remains at 99.20\%. The aggregate improvement is driven primarily by the Goal suite, alongside high success on object and spatial tasks. Its 99.20\% overall success matches SRPO, while retaining at least 98.40\% success in every suite.

\paragraph{Robustness across distribution shifts.}
We next evaluate whether this high standard-task performance extends to changes in camera viewpoint, robot initial state, language, lighting, background, visual noise, and scene layout using LIBERO-Plus. EffectWAM Small succeeds in 9,338 of 10,030 episodes, yielding 93.10\% overall success (Table~\ref{tab:libero_plus}). The rate is computed over all episodes to account for unequal category sizes. The 93.10\% overall rate is 0.30 points above the reported InternW0-$\Delta$ result. Category-level results show how this performance extends across distinct sources of variation.

\begin{table}[tbp]
\centering
\caption[LIBERO-Plus success rates (\%). Overall aggregates success over evaluation episodes.]{%
LIBERO-Plus success rates (\%). Overall aggregates success over evaluation episodes.
The highest score is shown in
\textcolor{red}{\textbf{bold red}}, and the second-highest distinct score
is \underline{underlined}.}
\label{tab:libero_plus}
\small
\setlength{\tabcolsep}{4pt}
\begin{adjustbox}{max width=\linewidth}
\begin{tabular}{@{}lccccccccc@{}}
\toprule
Method & Size (B) & Camera & Robot & Language & Light & Bkgd & Noise & Layout & Overall \\
\midrule
SRPO (aug.)~\cite{fei2025srpo} & 7.0 & 83.4 & 62.0 & 73.6 & \textcolor{red}{\textbf{97.2}} & 97.7 & 85.7 & 75.2 & 82.1 \\
OpenVLA-OFT~\cite{kim2025fine} & 7.0 & 56.4 & 31.9 & 79.5 & 88.7 & 93.3 & 75.8 & 74.2 & 69.6 \\
RIPT-VLA~\cite{tan2025riptvla} & 7.0 & 55.2 & 31.2 & 77.6 & 88.4 & 91.6 & 73.5 & 74.2 & 68.4 \\
NORA-Long~\cite{hung2025nora} & 3.0 & 2.20 & 37.0 & 65.1 & 45.7 & 58.6 & 12.8 & 62.1 & 39.0 \\
$\pi_0$~\cite{black2026pi0visionlanguageactionflowmodel} & 3.3 & 13.8 & 6.0 & 58.8 & 85.0 & 81.4 & 79.0 & 68.9 & 53.6 \\
$\pi_0$-FAST~\cite{pertsch2025fast} & 3.0 & 65.1 & 21.6 & 61.0 & 73.2 & 73.2 & 74.4 & 68.8 & 61.6 \\
InternW0-$\Delta$~\cite{miao2026internw0delta} & 8.0 & \underline{90.6} & \underline{91.1} & \underline{92.9} & 95.8 & \underline{99.4} & \textcolor{red}{\textbf{94.0}} & \underline{88.4} & \underline{92.8} \\
\textbf{EffectWAM Small} & 8.0 & \textcolor{red}{\textbf{91.7}} & \textcolor{red}{\textbf{91.4}} & \textcolor{red}{\textbf{93.4}} & \underline{96.5} & \textcolor{red}{\textbf{99.6}} & \underline{93.4} & \textcolor{red}{\textbf{88.7}} & \textcolor{red}{\textbf{93.1}} \\
\bottomrule
\end{tabular}
\end{adjustbox}
\par\vspace{3pt}{\footnotesize SRPO (aug.) uses the augmented-data configuration. Evaluations of OpenVLA-OFT, RIPT-VLA, NORA, and the $\pi_0$ variants are from LIBERO-Plus~\cite{fei2026liberoplus}.}
\end{table}

High success extends across the seven perturbation categories: six exceed 91\%, and all reach at least 88.66\%. Performance is particularly strong under background and lighting variation, at 99.54\% and 96.50\%, respectively, while language and visual-noise variations both exceed 93\%. Camera and Robot variations also retain success above 91\%. These results show that high task completion extends across both appearance changes and changes to the camera viewpoint or robot initial state, complementing the standard-task results with broad perturbation coverage.

\paragraph{Bimanual manipulation under randomization.}
RoboTwin 2.0 tests whether the policy maintains bimanual manipulation performance when the same task set is evaluated under scene randomization. We use 50 tasks with 100 episodes per task in each of the Clean and Randomized settings, totaling 10,000 episodes. EffectWAM Small achieves 91.82\% success on Clean and 74.22\% on Randomized, with an equal-weight average of 83.02\% (Table~\ref{tab:robotwin}). All methods in this comparison use only Clean demonstrations for benchmark adaptation. Randomized evaluation therefore tests transfer to scene conditions absent from those demonstrations.

\begin{table}[tbp]
\centering
\caption[RoboTwin 2.0 success rates (\%) under Clean-only training, evaluated on 50 tasks in both Clean and Randomized test conditions.]{%
RoboTwin 2.0 success rates (\%) under Clean-only training, evaluated on 50 tasks in both Clean and Randomized test conditions.
The highest score is shown in
\textcolor{red}{\textbf{bold red}}, and the second-highest distinct score
is \underline{underlined}.}
\label{tab:robotwin}
\small
\setlength{\tabcolsep}{4pt}
\begin{adjustbox}{max width=\linewidth}
\begin{tabular}{@{}lcccc@{}}
\toprule
Method & Size (B) & Clean\(\to\)Clean & Clean\(\to\)Random & Average \\
\midrule
Fast-WAM~\cite{yuan2026fastwam} & 11.7 & 77.8 & 1.9 & 39.9 \\
4D-WAM~\cite{yang2026fourdwam} & 11.0 & 81.5 & 41.8 & 61.7 \\
OpenWAM-$\alpha$~\cite{wang2026openwam} & 11.7 & 89.4 & 48.7 & 69.0 \\
InternW0-$\Delta$~\cite{miao2026internw0delta} & 8.0 & 90.0 & 71.9 & 81.0 \\
PICA-WAM~\cite{zou2026picawam} & 8.0 & \underline{90.7} & \textcolor{red}{\textbf{74.7}} & \underline{82.7} \\
\midrule
\textbf{EffectWAM Small} & 8.0 & \textcolor{red}{\textbf{91.8}} & \underline{74.3} & \textcolor{red}{\textbf{83.1}} \\
\bottomrule
\end{tabular}
\end{adjustbox}
\par\vspace{3pt}{\footnotesize Results for Fast-WAM and 4D-WAM are from InternW0-$\Delta$~\cite{miao2026internw0delta}. 4D-WAM uses the LingBot-VA architecture~\cite{li2026lingbotva}.}
\end{table}

EffectWAM exceeds the reported InternW0-$\Delta$ results in both settings, by 1.82 percentage points on Clean and 2.32 points on Randomized. The advantage over this reference therefore extends to randomized bimanual manipulation, with a larger margin in the Randomized setting. Its 83.02\% average also exceeds PICA-WAM's reported 82.7\%, with a higher Clean success rate. Maintaining 74.22\% success under randomization after Clean-only training demonstrates transfer beyond the benchmark training conditions.

Individual tasks further illustrate stable execution across settings. For example, \texttt{shake\_bottle} maintains 100\% Clean and 99\% Randomized success. The nearly unchanged completion rate on this task shows that successful execution can be preserved under scene randomization. Full per-task results are reported in Appendix~\ref{app:robotwin_results}.

\subsection{Compositional Manipulation and Capability Analysis}
\label{sec:compositional_manipulation}
\label{sec:robocasa}
\label{sec:robodojo}
\label{sec:contribution_analysis}

\paragraph{Compositional generalization.}
We use RoboCasa365 to examine whether improved task execution extends to combinations of behaviors, including combinations not seen during training. Its Atomic-Seen, Composite-Seen, and Composite-Unseen splits separate performance on atomic tasks from performance on familiar and unfamiliar compositions.

We evaluate version 1.0.1 on 50 tasks with 30 episodes per task. The 1,500 episodes comprise 540 Atomic-Seen, 480 Composite-Seen, and 480 Composite-Unseen episodes. InternW0-$\Delta$ uses the same task grouping and episode budget as EffectWAM Small. EffectWAM is trained on Human300 only; the Target training data are not used. Overall success is weighted by the number of episodes in each split. Table~\ref{tab:robocasa} compares performance on the three splits and their episode-weighted overall success.

\begin{table}[tbp]
\centering
\caption[RoboCasa365 success rates (\%) on the 50 evaluation tasks.]{%
RoboCasa365 success rates (\%) on the 50 evaluation tasks.
The highest score is shown in
\textcolor{red}{\textbf{bold red}}, and the second-highest distinct score
is \underline{underlined}.}
\label{tab:robocasa}
\small
\setlength{\tabcolsep}{4pt}
\begin{adjustbox}{max width=\linewidth}
\begin{tabular}{@{}lccccc@{}}
\toprule
Method & Size (B) & Atomic-Seen & Composite-Seen & Composite-Unseen & Overall \\
\midrule
Xiaomi-Robotics-1~\cite{xiaomi2026robotics1} & 5.1 & \textcolor{red}{\textbf{80.2}} & \textcolor{red}{\textbf{57.1}} & \textcolor{red}{\textbf{32.1}} & \textcolor{red}{\textbf{57.4}} \\
RLDX-1~\cite{kim2026rldx} & $\approx7$ & 67.6 & 27.9 & 8.5 & 36.0 \\
InternW0-$\Delta$~\cite{miao2026internw0delta}$^{\dagger}$ & 8.0 & 76.5 & 47.7 & 17.3 & 48.3 \\
\textbf{EffectWAM Small} & 8.0 & \underline{76.9} & \underline{48.2} & \underline{19.4} & \underline{49.3} \\
\bottomrule
\end{tabular}
\end{adjustbox}
\par\vspace{3pt}{\footnotesize $\dagger$: Evaluated by us.}
\end{table}

EffectWAM improves success across all three splits, with 739 successful episodes compared with 725 for InternW0-$\Delta$. Overall success increases from 48.33\% to 49.27\%, a gain of 0.93 percentage points based on the episode counts. The largest improvement occurs on Composite-Unseen: ten of the 14 additional successes come from this split, corresponding to a 2.08-point gain. Atomic-Seen and Composite-Seen each contribute two additional successes. The benefit therefore extends to unfamiliar task combinations, where the gain is larger than on the better-performing atomic tasks.

\paragraph{Capability-specific performance.}
We use RoboDojo to characterize EffectWAM's performance across Generalization, Precision, Long-Horizon, Memory, and Open capabilities. Success rate (SR) measures complete execution, while the benchmark's task scores also capture partial progress. Together with individual task results, these measures reveal the strongest capabilities and task behaviors of the policy.

We evaluate both configurations on all 42 RoboDojo simulation tasks: Small at training step 3,000 using seed 0, and Large at step 7,500 using three evaluation seeds. Each seed uses 25 standard and 25 randomized episodes for each of the 12 Generalization tasks and 50 episodes for each of the remaining 30 tasks. This yields 2,100 episodes for Small and 6,300 episodes for Large. Following the official protocol~\cite{chen2026robodojo}, we combine standard and randomized episodes for each Generalization task, average task metrics within each capability, and assign equal weight to the five capability means. Small results are single-seed estimates without an across-seed standard deviation; Large results report the mean and population standard deviation across seeds.

\begin{table}[tbp]
\centering
\caption[RoboDojo results at two model scales. Capability columns and Overall report success rates (\%); Score is the overall partial-credit task score. Results shown with $\pm$ are reported as the mean $\pm$ standard deviation across three seeds; EffectWAM Small uses one seed.]{%
RoboDojo results at two model scales. Capability columns and Overall report success rates (\%); Score is the overall partial-credit task score. Results shown with $\pm$ are reported as the mean $\pm$ standard deviation across three seeds; EffectWAM Small uses one seed.
The highest score is shown in
\textcolor{red}{\textbf{bold red}}, and the second-highest distinct score
is \underline{underlined}.}
\label{tab:robodojo}
\label{tab:robodojo_small}
\small
\setlength{\tabcolsep}{3pt}
\begin{adjustbox}{max width=\linewidth}
\begin{tabular}{@{}lcccccccc@{}}
\toprule
Method & Size (B) & Generalization & Precision & Long-Horizon & Memory & Open & Overall & Score \\
\midrule
\multicolumn{9}{@{}l}{\textit{Small scale}} \\
\addlinespace[2pt]
Awomo-0.5V~\cite{team2026awomo} & $\approx10.7$ & \underline{20.80} & \textcolor{red}{\textbf{28.10}} & \underline{25.00} & 12.50 & \textcolor{red}{\textbf{28.10}} & \underline{22.90} & \underline{28.70} \\
PrismWAM (Small)~\cite{xin2026prismwam} & $\approx11.7$ & 19.56 & 19.50 & 22.08 & \underline{30.11} & 1.500 & 18.55 & 25.10 \\
\textbf{EffectWAM Small} & $\approx8.40$ & \textcolor{red}{\textbf{24.33}} & \underline{20.75} & \textcolor{red}{\textbf{27.00}} & \textcolor{red}{\textbf{55.00}} & \underline{12.25} & \textcolor{red}{\textbf{27.87}} & \textcolor{red}{\textbf{33.53}} \\
\midrule
\multicolumn{9}{@{}l}{\textit{Large scale}} \\
\addlinespace[2pt]
VPP2~\cite{guo2026vpp2} & $\approx24.7$ & \underline{29.50} & \textcolor{red}{\textbf{35.17}} & \textcolor{red}{\textbf{37.00}} & \textcolor{red}{\textbf{55.56}} & 4.08 & \underline{32.26} & \underline{39.26} \\
PrismWAM (Large)~\cite{xin2026prismwam} & $\approx25.0$ & $29.33\pm0.98$ & $31.08\pm2.66$ & $31.00\pm0.74$ & $50.67\pm0.27$ & $\underline{4.42}\pm0.66$ & $29.30\pm0.61$ & $36.23\pm0.55$ \\
\textbf{EffectWAM Large} & $\approx26.7$ & $\textcolor{red}{\textbf{33.22}}\pm1.14$ & $\underline{34.67}\pm0.72$ & $\underline{36.92}\pm1.12$ & $\underline{54.22}\pm1.37$ &$\textcolor{red}{\textbf{11.00}}\pm1.24$ & $\textcolor{red}{\textbf{34.01}}\pm0.12$ & $\textcolor{red}{\textbf{40.29}}\pm0.32$ \\
\bottomrule
\end{tabular}
\end{adjustbox}
\par\vspace{3pt}{\footnotesize Sizes are estimated from model components.}
\end{table}

\begin{samepage}
Table~\ref{tab:robodojo} groups the comparisons by model scale, with a parameter range spanning less than 5B within each group. EffectWAM Small achieves 27.87\% overall success and a task score of 33.53. Its overall success rate exceeds the reported Awomo-0.5V and PrismWAM (Small) results. Memory is its strongest capability, with 55.00\% success and a task score of 55.38; its Memory success is also higher than the selected small-scale baselines. Long-Horizon reaches 27.00\% success and a score of 41.29, indicating credited progress beyond complete task execution. Precision and Open remain weaker than Awomo-0.5V. The single-seed evaluation does not quantify run-to-run variability.\par
\end{samepage}

At the larger scale, EffectWAM Large achieves an overall success rate of 34.01\% and a task score of 40.29, exceeding VPP2 by 1.75 and 1.03 points, respectively (Table~\ref{tab:robodojo}). Compared with PrismWAM$^{*}$ (Large), it improves overall success by 4.71 percentage points and the task score by 4.06 points, with higher success across all five capabilities. Relative to VPP2, the largest success-rate advantage is on Open tasks, where success increases from 4.08\% to 11.00\%; Generalization also improves from 29.50\% to 33.22\%. The advantage over VPP2 is therefore concentrated in the Open and Generalization capabilities. Task scores range from 0 to 100 and credit partial completion, whereas success requires full task completion. Within this evaluation, Memory is the strongest capability, with 54.22\% success and a closely aligned task score of 54.62. Individual tasks show particularly high completion rates: \texttt{press\_by\_number} succeeds in 149 of 150 episodes and \texttt{cover\_blocks} in 144 of 150. The close agreement between the category's two metrics indicates that its score is dominated by complete task execution, with these selected tasks providing concrete examples of reliable performance.

Long-Horizon achieves a task score of 53.44 alongside 36.92\% full-task success. The task score captures additional credited progress beyond the episodes counted as complete successes, providing a complementary view of the policy's execution on extended tasks. Together, the Memory and Long-Horizon results highlight two aspects of EffectWAM Large's capability profile: frequent complete execution on Memory tasks and substantial credited progress on Long-Horizon tasks.

\input{iFlyBot_sections/execution_visualization}

\input{iFlyBot_sections/attention_visualization}

The experiments highlight high standard-task success, broad perturbation coverage, and improved compositional execution. EffectWAM Small achieves 99.20\% success on LIBERO and 93.10\% on LIBERO-Plus, exceeds the reported InternW0-$\Delta$ results in both RoboTwin settings, and improves RoboCasa365 success across all three splits. RoboCasa365 gains are largest on unseen compositions. On RoboDojo, Small reaches 27.87\% overall success with strong Memory performance, and Large reaches 34.01\%.

%% file: iFlyBot_sections/execution_visualization.tex
\begin{figure}[!t]
    \centering
    \includegraphics[width=\linewidth]{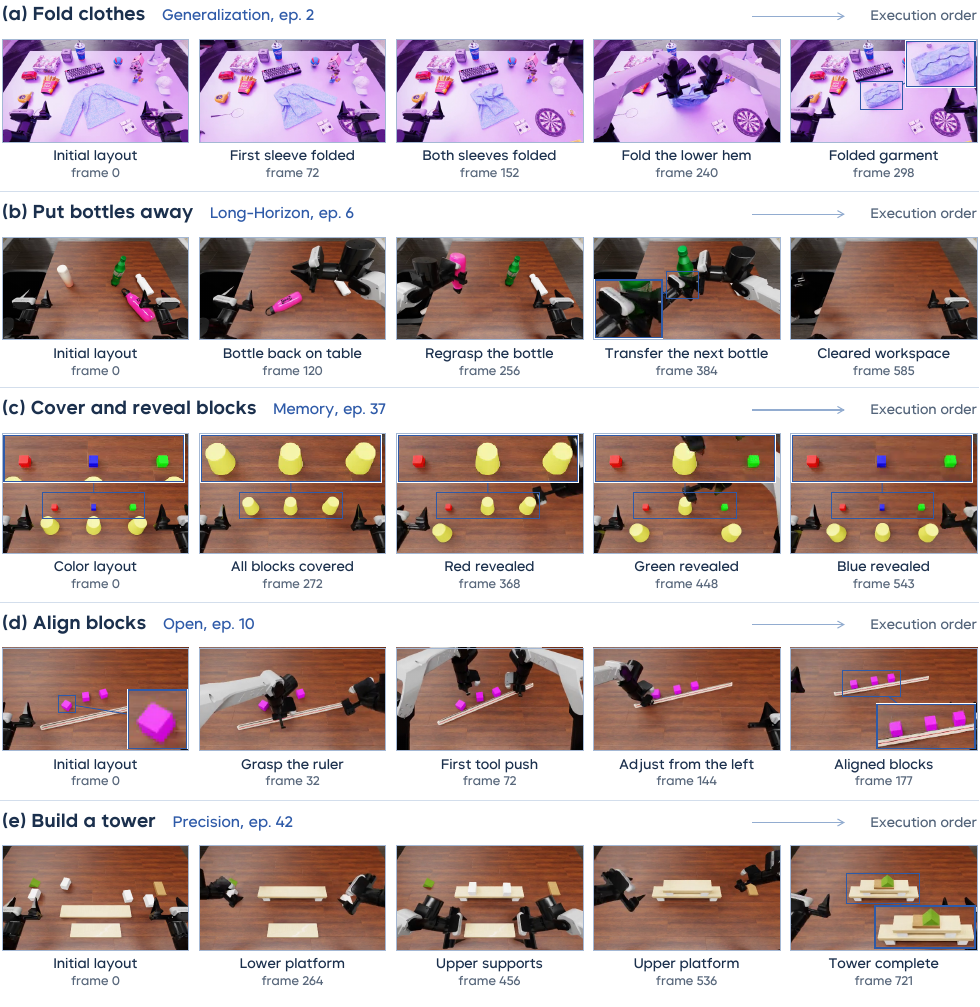}
    \caption{Selected successful RoboDojo simulation executions of EffectWAM Large (26B): (a) folding clothes in a randomized scene, (b) putting bottles into a dustbin, (c) covering and revealing colored blocks, (d) aligning blocks with a ruler, and (e) building a tower. Each row follows one recorded head-camera trajectory from left to right. Frames capture task-relevant events at unequal intervals; episode IDs and zero-based source-video frame indices identify the recordings. Main views use a fixed crop within each row. Blue outlines locate the regions shown in blue-bordered insets from the same frame; the Memory detail crop is fixed across stages.}
    \label{fig:robodojo_execution}
\end{figure}

\paragraph{Execution visualization.}
To relate the capability scores to concrete manipulation behavior, Fig.~\ref{fig:robodojo_execution} presents successful executions of EffectWAM Large (26B) from each of RoboDojo's five capability categories~\cite{chen2026robodojo}. The Generalization example folds the sleeves and lower hem of a rotated garment in a cluttered scene with colored lighting, combining deformable-object handling with changes in visual appearance. In the Long-Horizon sequence, the robot disposes of four bottles with different initial poses. A bottle is regrasped after returning to the tabletop, and later bottles are passed between the grippers before disposal, maintaining progress across repeated grasping and transfer operations.

The Memory example first covers three colored blocks, then reveals them in the required red--green--blue order. This differs from their spatial order, so the color--position associations observed before occlusion remain relevant when selecting which cup to lift. In Open, the robot uses a ruler to bring the blocks into line, then repositions it from the left end to correct the remaining misalignment. The Precision example alternates individual support-block placements with coordinated transport of the horizontal platforms, then adds the two top pieces to complete an eight-piece tower. Each new layer depends on the support arrangement below it, and the structure remains standing after release. These sequences complement the aggregate results in Table~\ref{tab:robodojo} with examples of the interactions required for completion. Appendix~\ref{app:robodojo_executions} presents 25 additional successful trajectories across the same five categories.

%% file: iFlyBot_sections/attention_visualization.tex
\begin{figure}[!htbp]
    \centering
    \includegraphics[width=\linewidth]{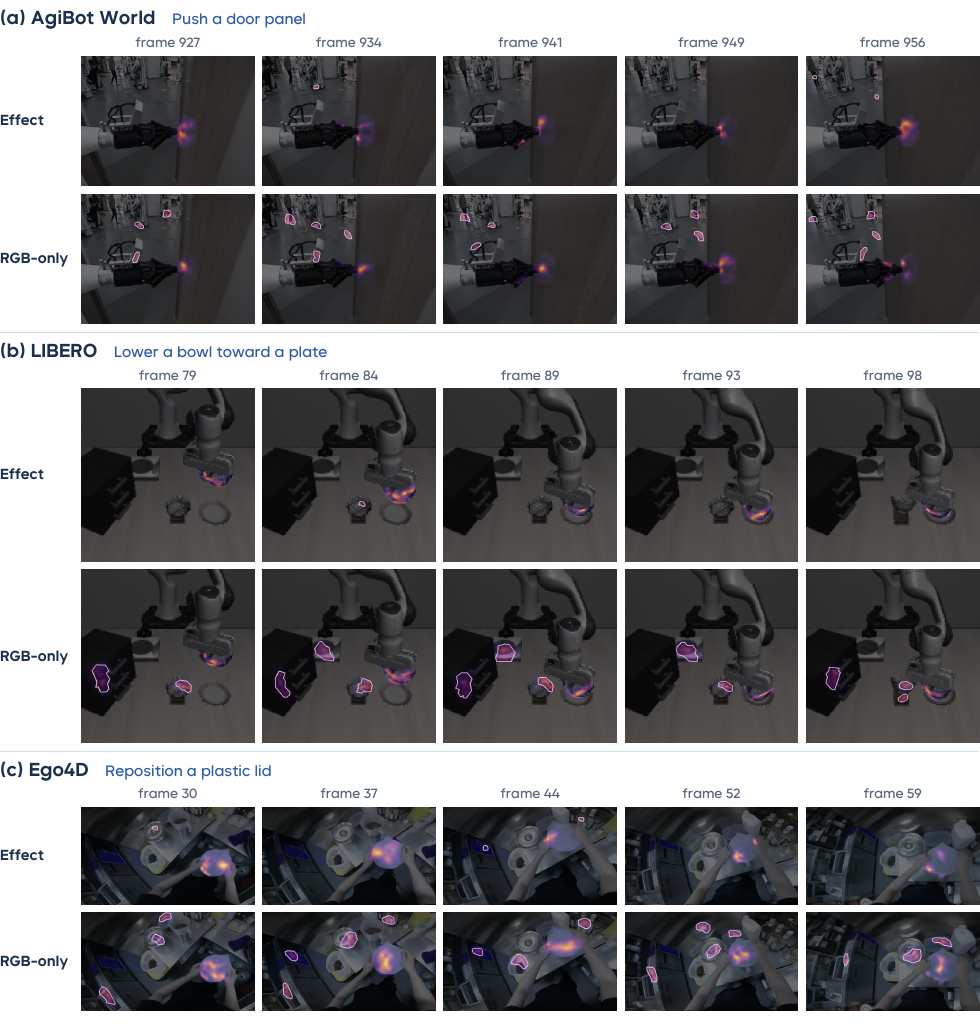}
    \caption{Qualitative attention comparison for EffectWAM Small (8B): (a) pushing a door panel, (b) lowering a bowl toward a plate, and (c) repositioning a plastic lid. Effect denotes attention for future Effect prediction in the full model; RGB-only denotes attention for future-RGB prediction without Effect supervision. Maps aggregate attention from the final eight layers to historical input frames, averaging over heads and summing over future-prediction query tokens, with the same normalization and color scale for both variants. Paired rows show the same frames in temporal order, with zero-based source-frame indices above each column; the sampled frames are not consecutive. Brighter colors indicate stronger displayed attention. Light pink polygon outlines highlight background regions with off-target attention for easier visual comparison. These outlines are added annotations; attention strength is represented by the underlying heatmaps.}
    \label{fig:attention_comparison}
\end{figure}

\paragraph{Attention visualization.}
To examine the visual information emphasized during prediction, Fig.~\ref{fig:attention_comparison} compares attention for future Effect prediction in the full model with future-RGB attention after removing Effect supervision. Both use the Small (8B) configuration. In AgiBot World~\cite{bu2025agibotworld}, the Effect response remains near the gripper--door interaction during pushing, with sparse, faint responses elsewhere; the RGB maps also highlight distant people, nearby fixtures, and the floor region. In LIBERO~\cite{liu2023libero}, both variants attend to the held bowl as it descends toward the plate. The RGB-only maps additionally emphasize non-target bowls and the static cabinet on the left. In Ego4D~\cite{grauman2022ego4d}, the held plastic lid remains the main focus of the Effect maps as the first-person viewpoint changes, while RGB prediction also highlights the metal lid on the counter, containers, and peripheral equipment. Across these examples, Effect has fewer, smaller, and generally dimmer off-target responses, while retaining prominent activity near the manipulated object. This pattern is consistent with the intended emphasis on action-related visual change and provides a qualitative complement to the robustness results.

\FloatBarrier

%% file: iFlyBot_sections/conclusions.tex
\section{Limitations and Conclusion}
\label{sec:conclusion}

\textbf{Limitations.}
Performance remains uneven across evaluation conditions. The decrease under RoboTwin randomization, the low success on unseen Robocasa365 compositions, and the difficulty of RoboDojo's Open tasks identify robustness and generalization as continuing challenges. On long-horizon RoboDojo tasks, the gap between partial-credit scores and full-task success further highlights the need to sustain useful behavior through task completion.

The method also depends on the availability of geometric and physical supervision. CaPE requires reliable camera extrinsics; uncalibrated trajectories use the shared pose-free route. Action Grounding supervises only available, consistently aligned measurements, limiting its coverage of physical quantities across heterogeneous data sources. At inference, optional video generation keeps video synthesis outside the control path, but joint Effect--action sampling still incurs computational cost. Prediction horizons remain configured inputs and do not automatically adapt to subgoal duration.

\textbf{Conclusion.}
We presented \textbf{EffectWAM}, a world--action model that connects visual change prediction with executable robot control through Effect, CaPE, and Action Grounding. Effect encodes visual change as endpoint differences relative to a shared current-observation anchor. CaPE incorporates relative camera geometry into endpoint pooling when calibration is available, while Action Grounding aligns the representation with measured motion. A shared generation expert jointly predicts Effect and actions through flow matching and reuses the sampled pair for optional video forecasting. Training on more than 20,000 hours of heterogeneous embodied data brings these components together across robot and human demonstrations with different action representations.

Evaluation across five manipulation benchmarks demonstrates strong standard-task performance and improved success over the original policy on Robocasa365, with the largest gain on unseen compositions. Cross-baseline evaluations further indicate that Action Grounding improves performance across the evaluated baseline policies, supporting its role as a transferable training signal. The remaining sensitivity to environmental variation and difficulty in completing composite tasks motivate further work on robust physical grounding, adaptive action horizons, and more efficient joint sampling.

%% file: iFlyBot_sections/appendix_results.tex
\appendix
\section{Pretraining Data Statistics}
\label{app:data_inventory}
The pretraining data comprise 34 sources with a total recorded duration of 25,292.8 hours. Table~\ref{tab:training_data_statistics} provides the per-source durations underlying the data composition in Fig.~\ref{fig:dataset_composition}. Unused chain-of-thought data are excluded, and alternative processing versions of the same source are counted only once.

\begin{table}[htbp]
\centering
\caption{Recorded duration of each pretraining data source, in hours.}
\label{tab:training_data_statistics}
\small
\begin{tabular}{@{}lc@{}}
\toprule
Source & Hours \\
\midrule
Builddotai Egocentric-10K & 9,984.1 \\
HY-Embodied-0.5 & 2,158.8 \\
HiFi-UMI-2K & 2,131.3 \\
ABC-130K & 1,664.9 \\
DROID & 1,276.1 \\
Jianzhi v1 & 1,156.5 \\
Mifeng batch2 & 994.8 \\
Lightwheel EgoProStandard & 994.3 \\
InternData M1 Franka & 848.3 \\
InternData A1 & 820.3 \\
RH20T & 784.7 \\
EgoDex v1 & 724.2 \\
Fractal & 350.2 \\
RoboCOIN & 215.6 \\
BC-Z & 165.8 \\
GM100 & 151.9 \\
BitRobot HIW-500 & 127.9 \\
Mifeng batch1 & 111.6 \\
FurnitureBench & 109.7 \\
EgoVerse Lightwheel part1 & 100.4 \\
EgoVerse Aria v1 & 74.6 \\
InternData M1 AgileX & 62.5 \\
PineLab ManuFacet & 49.3 \\
EgoVerse Mecka v1 & 38.3 \\
Imitator robot & 37.0 \\
FineArt 90Hz & 33.3 \\
Imitator sim & 30.9 \\
RoboMIND2 AgileX & 28.4 \\
RoboDojo (simulation) & 20.5 \\
EgoVerse Lightwheel part2 & 12.6 \\
GPT-Astra (robot operation) & 10.5 \\
Stanford Hydra v2 & 10.0 \\
Imitator human v1 & 8.1 \\
Berkeley UR5 & 5.4 \\
\midrule
Total & 25,292.8 \\
\bottomrule
\end{tabular}
\end{table}
\clearpage
\section{Per-Task Results on RoboTwin 2.0}
\label{app:robotwin_results}
\begin{table}[!htbp]
\centering
\caption{Per-task RoboTwin 2.0 success rates (\%) for EffectWAM Small (8B).}
\scriptsize
\setlength{\tabcolsep}{5pt}
\begin{tabular}{@{}lcclcc@{}}
\toprule
Task & Clean\(\to\)Clean & Clean\(\to\)Random & Task & Clean\(\to\)Clean & Clean\(\to\)Random \\
\midrule
adjust\_bottle & 100 & 96 & place\_container\_plate & 99 & 80 \\
beat\_block\_hammer & 99 & 74 & place\_dual\_shoes & 92 & 61 \\
blocks\_ranking\_rgb & 91 & 85 & place\_empty\_cup & 100 & 75 \\
blocks\_ranking\_size & 71 & 42 & place\_fan & 96 & 77 \\
click\_alarmclock & 100 & 90 & place\_burger\_fries & 100 & 98 \\
click\_bell & 100 & 91 & place\_mouse\_pad & 91 & 71 \\
dump\_bin\_bigbin & 99 & 94 & place\_object\_basket & 87 & 64 \\
grab\_roller & 100 & 77 & place\_object\_scale & 98 & 83 \\
handover\_block & 93 & 67 & place\_object\_stand & 98 & 87 \\
handover\_mic & 99 & 82 & place\_phone\_stand & 92 & 64 \\
lift\_pot & 100 & 81 & move\_pillbottle\_pad & 100 & 71 \\
move\_can\_pot & 94 & 65 & place\_shoe & 96 & 80 \\
move\_playingcard\_away & 98 & 73 & press\_stapler & 91 & 80 \\
move\_stapler\_pad & 74 & 65 & put\_bottles\_dustbin & 93 & 88 \\
hanging\_mug & 56 & 40 & put\_object\_cabinet & 67 & 38 \\
open\_laptop & 94 & 78 & rotate\_qrcode & 93 & 48 \\
open\_microwave & 68 & 43 & scan\_object & 90 & 59 \\
pick\_diverse\_bottles & 91 & 78 & shake\_bottle & 100 & 99 \\
pick\_dual\_bottles & 100 & 81 & shake\_bottle\_horizontally & 100 & 97 \\
place\_a2b\_left & 92 & 70 & stack\_blocks\_three & 99 & 81 \\
place\_a2b\_right & 94 & 64 & stack\_blocks\_two & 100 & 89 \\
place\_bread\_basket & 93 & 91 & stack\_bowls\_three & 75 & 70 \\
place\_bread\_skillet & 96 & 78 & stack\_bowls\_two & 88 & 77 \\
place\_can\_basket & 75 & 67 & stamp\_seal & 94 & 65 \\
place\_cans\_plasticbox & 98 & 93 & turn\_switch & 77 & 44 \\
\bottomrule
\end{tabular}
\end{table}

\input{iFlyBot_sections/appendix_execution_visualization}

%% file: iFlyBot_sections/appendix_execution_visualization.tex
\clearpage
\section{Additional RoboDojo Executions}
\label{app:robodojo_executions}
\begin{figure}[!htbp]
    \centering
    \includegraphics[width=\linewidth]{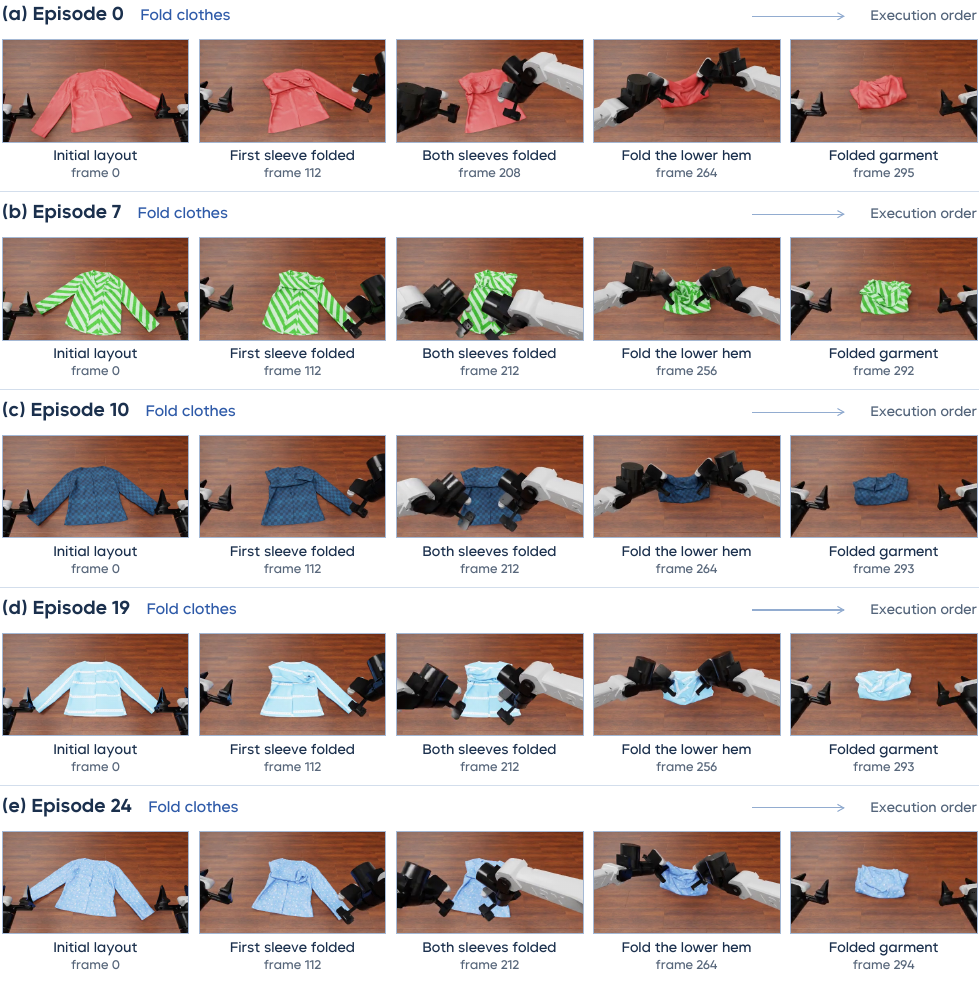}
    \caption{Additional Generalization executions on \texttt{fold\_clothes}, episodes 0, 7, 10, 19, and 24. Across different garment appearances, the robot folds the sleeves inward before using both grippers to fold the lower hem. Intermediate frames show the two sleeve folds; the last column shows the released garment.}
    \label{fig:robodojo_extra_generalization_standard}
\end{figure}

\begin{figure}[!htbp]
    \centering
    \includegraphics[width=\linewidth]{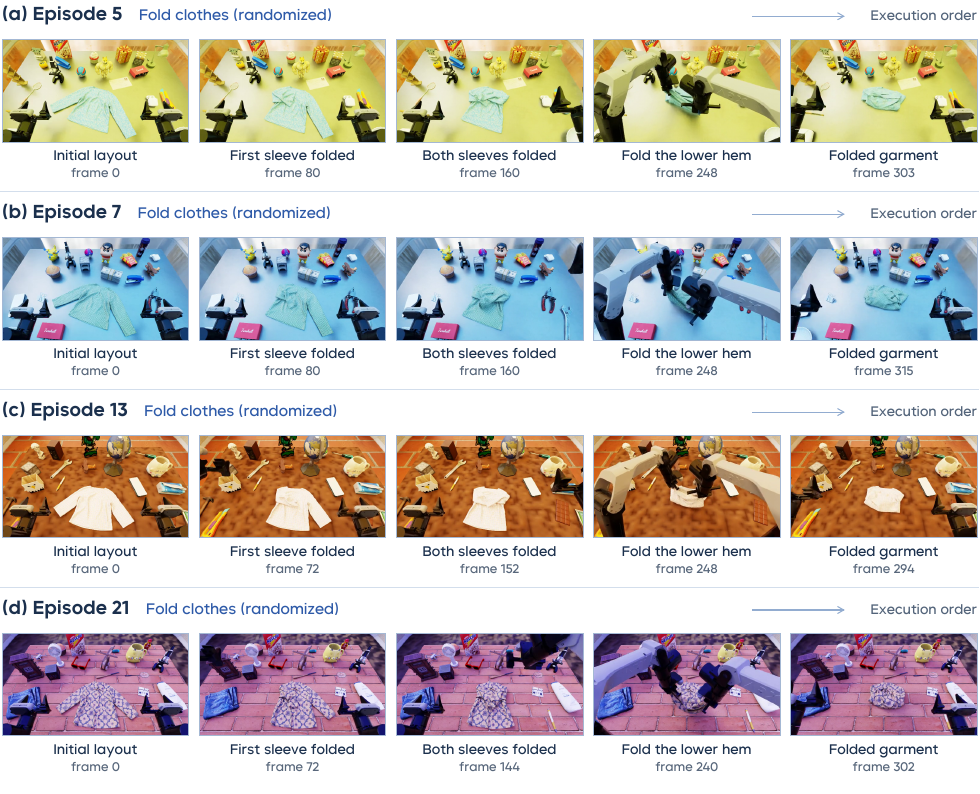}
    \caption{Additional Generalization executions on \texttt{fold\_clothes\_random}, episodes 5, 7, 13, and 21. The workspace crops retain the surrounding objects and visual conditions, rather than isolating the garment. Each row follows one episode in temporal order, with the final frame showing the fold after release.}
    \label{fig:robodojo_extra_generalization_random}
\end{figure}

\begin{figure}[!htbp]
    \centering
    \includegraphics[width=\linewidth]{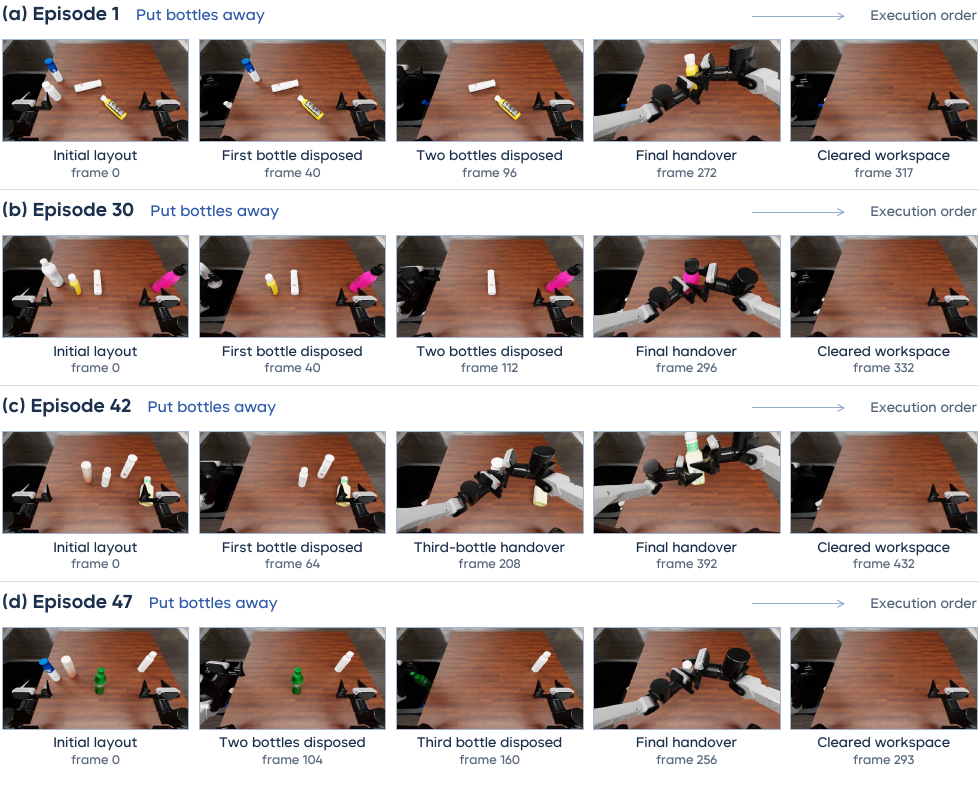}
    \caption{Additional Long-Horizon executions on \texttt{put\_bottles\_into\_dustbin}, episodes 1, 30, 42, and 47. Event frames show disposal progress and inter-arm handovers. The terminal view shows the cleared tabletop after the four-bottle sequence; each row retains a fixed crop that includes the dustbin on the left.}
    \label{fig:robodojo_extra_long_horizon}
\end{figure}

\begin{figure}[!htbp]
    \centering
    \includegraphics[width=\linewidth]{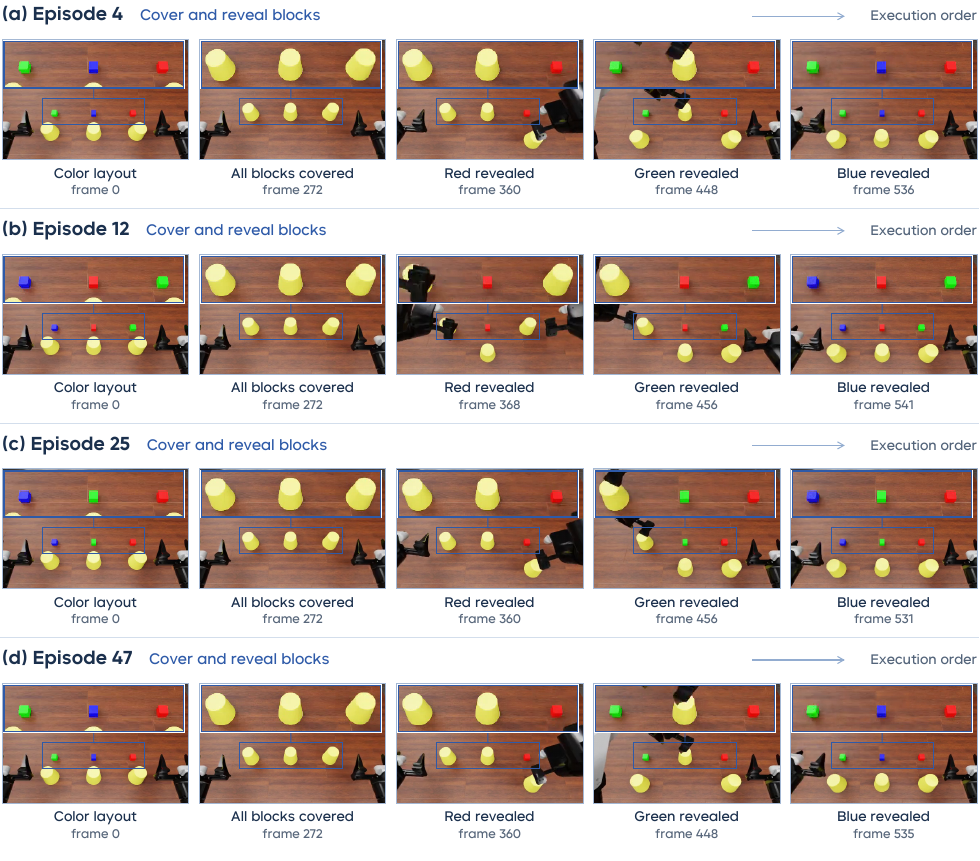}
    \caption{Additional Memory executions on \texttt{cover\_blocks}, episodes 4, 12, 25, and 47. Columns show the initial colors, complete occlusion, and the states after revealing red, green, and blue. The uncovering order follows color rather than a fixed left-to-right sweep; the final frame also shows the cups returned away from the blocks.}
    \label{fig:robodojo_extra_memory}
\end{figure}

\begin{figure}[!htbp]
    \centering
    \includegraphics[width=\linewidth]{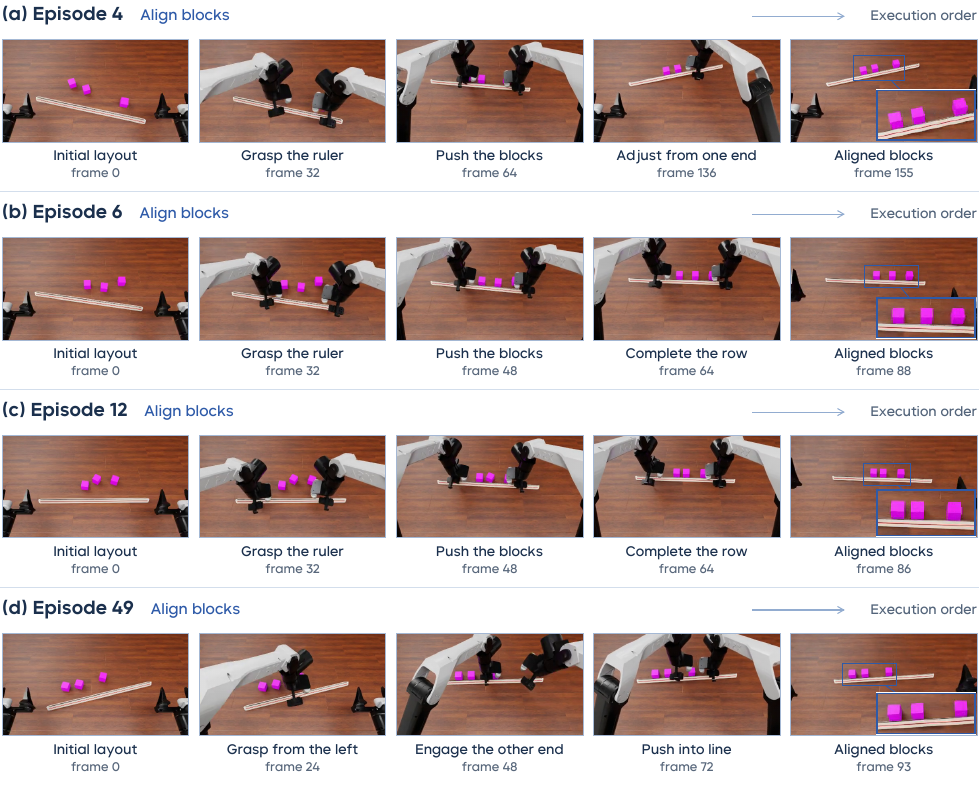}
    \caption{Additional Open executions on \texttt{align\_blocks}, episodes 4, 6, 12, and 49. The event frames retain the initial arrangement, tool engagement, pushing or adjustment, and final alignment. Blue-bordered details magnify the terminal blocks and their relation to the ruler.}
    \label{fig:robodojo_extra_open}
\end{figure}

\begin{figure}[!htbp]
    \centering
    \includegraphics[width=\linewidth]{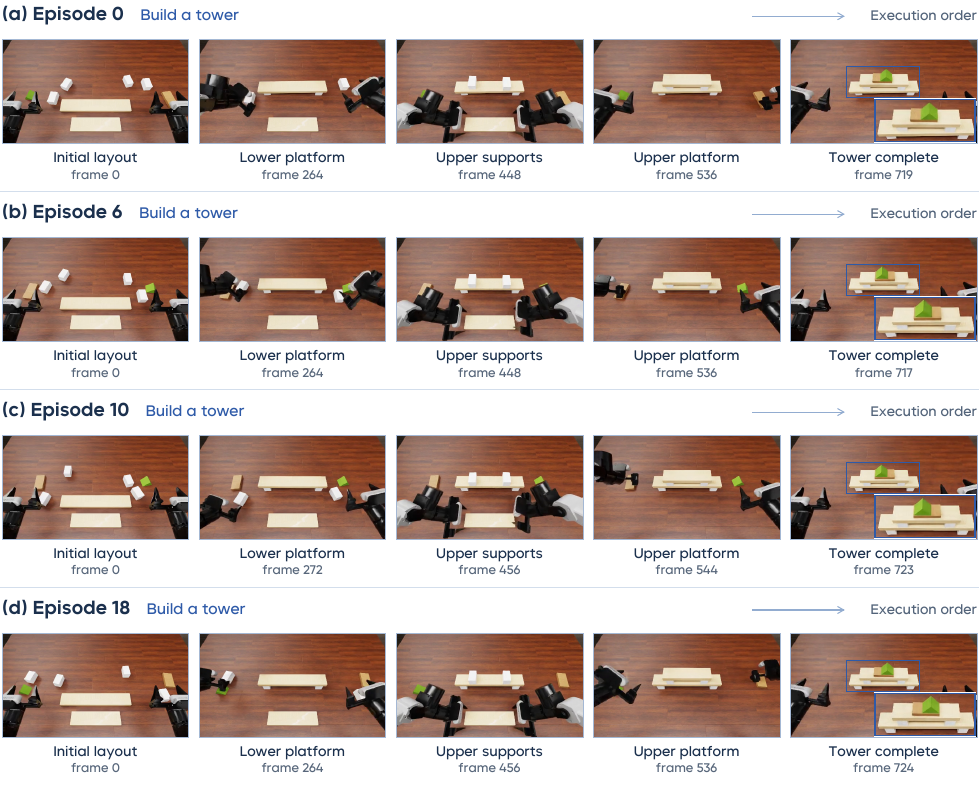}
    \caption{Additional Precision executions on \texttt{build\_tower}, episodes 0, 6, 10, and 18. Each sequence assembles four support blocks, two horizontal platforms, and two top pieces. The intermediate frames show successive support levels, and the final detail highlights the completed eight-piece structure.}
    \label{fig:robodojo_extra_precision}
\end{figure}

Figures~\ref{fig:robodojo_extra_generalization_standard}--\ref{fig:robodojo_extra_precision} present 25 additional successful executions of EffectWAM Large (26B), complementing the five examples in Fig.~\ref{fig:robodojo_execution}. The sequences follow RoboDojo's capability grouping and include both standard and randomized clothing-folding tasks. Each row is a separate recorded episode; the selected frames preserve the initial state, intermediate interactions, and completion. Indices refer to zero-based source-video frames, with unequal intervals between the selected events. Blue-bordered details are taken from the outlined regions of the same frames.

\clearpage

\paragraph{Generalization under scene variation.}
The randomized folding episodes change the garment pose and surrounding appearance while retaining the same folding objective. Figure~\ref{fig:robodojo_extra_generalization_random} includes colored illumination, varied tabletop textures, and nearby objects. The grasp locations therefore change with the garment, while the sequence still proceeds through the two sleeve folds and the final fold of the lower hem.

\clearpage

\paragraph{Progress across repeated transfers.}
The bottle-disposal task requires all four bottles to reach the dustbin. The episodes in Fig.~\ref{fig:robodojo_extra_long_horizon} vary the bottle shapes, positions, and initial poses. Bottles near the bin can be transferred directly, whereas bottles on the other side of the workspace are passed between grippers. The intermediate frames track this reduction in the remaining objects, including the handover of the final bottle before the workspace is cleared.

\clearpage

\paragraph{Color order after occlusion.}
In Fig.~\ref{fig:robodojo_extra_memory}, the initial color arrangements differ, but the robot first covers the blocks from left to right and subsequently reveals red, green, and blue in that order. The all-covered frame removes the visible color cues. Subsequent frames show the resulting spatially different uncovering sequences, making the dependence on the earlier color--position arrangement explicit. The detail windows use the same image region throughout each row so that the visible blocks and covered locations can be compared directly.

\clearpage

\paragraph{Alignment through tool contact.}
Figure~\ref{fig:robodojo_extra_open} shows different initial ruler orientations and block offsets. The ruler provides a shared contact surface for bringing the blocks into line without picking them up individually. Episode 4 includes an additional adjustment from one end of the ruler, while the other three episodes reach the aligned configuration in a continuous pushing sequence. The enlarged terminal views make the final block arrangement easier to inspect.

\clearpage

\paragraph{Building successive support layers.}
In Fig.~\ref{fig:robodojo_extra_precision}, the arms place the four white supports individually and transport the horizontal platforms together. Their positions must support the next layer as the structure grows. The locations of the two top pieces also vary: the arm used for each of these final placements changes between the episodes. The selected intermediate states expose the completed lower platform, the upper supports, and the upper platform; the last frame shows the full tower after the grippers withdraw.

\FloatBarrier